# GENERALIZED JERSEY NUMBER RECOGNITION USING MULTI-TASK LEARNING WITH ORIENTATION-GUIDED WEIGHT REFINEMENT

**Yung-Hui Lin**
Department of Electrical Engineering
Yuan Ze University
Taiwan
jennifer19990920@gmail.com

**Yu-Wen Chang**
Department of Electrical Engineering
Yuan Ze University
Taiwan
tony506672558@gmail.com

**Huang-Chia Shih**
Department of Electrical Engineering
Yuan Ze University
Taiwan
hcshih@saturn.yzu.edu.tw

**Takahiro Ogawa**
Faculty of Information Science and Technology
Hokkaido University
Japan
ogawa@lmd.ist.hokudai.ac.jp

May 27, 2024

## ABSTRACT

Jersey number recognition (JNR) has always been an important task in sports analytics. Improving recognition accuracy remains an ongoing challenge because images are subject to blurring, occlusion, deformity, and low resolution. Recent research has addressed these problems using number localization and optical character recognition. Some approaches apply player identification schemes to image sequences, ignoring the impact of human body rotation angles on jersey digit identification. Accurately predicting the number of jersey digits by using a multi-task scheme to recognize each individual digit enables more robust results. Based on the above considerations, this paper proposes a multi-task learning method called the angle-digit refine scheme (ADRS), which combines human body orientation angles and digit number clues to recognize athletic jersey numbers. Based on our experimental results, our approach increases inference information, significantly improving prediction accuracy. Compared to state-of-the-art methods, which can only handle a single type of sport, the proposed method produces a more diverse and practical JNR application. The incorporation of diverse types of team sports such as soccer, football, basketball, volleyball, and baseball into our dataset contributes greatly to generalized JNR in sports analytics. Our accuracy achieves 64.07% on Top-1 and 89.97% on Top-2, with corresponding F1 scores of 67.46% and 90.64%, respectively.

## 1 Introduction

One of the most important tasks of sports analytics is to show athletes their weaknesses and suggest tactics for improvement. This also allows coaches to devise latent game strategies. However, a systematic understanding of each player on the field is exceedingly difficult, requiring not only diverse sports data but also accurate player identification. Furthermore, due to the rapid movement of players and the distant placement of cameras on the field, player images typically have relatively low resolution with blurring. Therefore, facial recognition is impractical for identifying players during sports activities. Thus, jersey numbers have become a unique and crucial type of information for player identification. Correctly distinguishing the number of digits on a jersey image is also a critical issue for JNR. JNR cannot utilize natural image character recognition techniques directly because the precise position of the jersey number must be located. Even with the application of state-of-the-art text or object detectors, there is still a high probability of missing the target number location.

JNR is a challenging task in practice. In past research, JNR was generally regarded as a single task process [1, 2, 3, 4, 5]. Some published methods treated jersey number recognition as an overall classification task [1], [5]. In such a classification task, jersey numbers are typically divided into 100 categories (i.e., the numbers 0-99) along with an extra category to deal with unrecognizable cases. The disadvantage of using classification task methods is the prohibitive number of categories. Moreover, the number of images available for each learnable category is relatively low. For example, with respect to jersey number 93, the system can only learn that the number is 93 based on this method. However, if the number 93 is regarded as individual digits, it can learn that the tens-digit is 9, the ones-digit is 3, and the tens-digits from 90 to 99 are all 9. Such a model can learn repeatedly and achieve better results for certain rarely used numbers. The disadvantage of using models that consider the task as individual digits [2, 3, 4] is that it may be difficult to clearly distinguish the digits. Separating tens-digits from ones-digits is also a highly challenging task. Vats et al. [6] addressed the problem of jersey number recognition from a multi-task perspective. They divided jersey numbers into overall and individual number processes, but ignored the accuracy of digit interpretation and the rotation of the human body. To improve recognition accuracy, the present study proposes a multi-task learning approach with a fine-tuning scheme that considers human body orientation angle and digit number, allowing the digit number to support the learning of individual numbers and successfully improve prediction accuracy.

The main contributions of this paper are as follows:

1 ) This study presents a more general JNR method for diverse types of sports instead of focusing on a single sport. This enables the model to be used in broader applications and enhances its generalization ability.

2 ) We combine human body posture and orientation angle into a number-of-digits estimation to assist and improve the accuracy of the model.

3 ) A novel model is proposed that combines multi-task and digit-wise learning within an architecture that effectively improves model prediction accuracy. Compared with conventional multi-task learning [6], the present study achieves improvements of 10.81% and 11.38% on Top-1 and Top-2 accuracy, respectively.

4 ) We have compiled a new dataset that combines the SoccerNet dataset [7] with jersey number images from diverse team sports videos, including football, basketball, volleyball, and baseball videos.

## 2   Related Work

The field of sports analytics, particularly with respect to computer vision, has experienced significant advancements facilitated by deep learning techniques. A large number of studies [1, 2, 3, 4, 5], [8, 9, 10, 11] address various aspects of player identification or recognition by detecting and recognizing players and their jersey numbers in team sport videos.

### 2.1   Player Identification Using Facial Recognition

Conventionally, facial features have been used as a crucial identifier for player recognition [12, 13, 14]. The AdaBoost algorithm can be used to detect individual players and their facial features [12]. Similarly, an AdaBoost algorithm with Haar-like features can be employed for feature selection and classification, and also utilized for facial recognition [13]. Ballan et al. [14] integrated a face detection algorithm into the AdaBoost detector. However, the availability of facial images captured from large sports fields is limited, so it is impractical to rely solely on facial features for player identification.

### 2.2   Improved Player Identification Using Deep Learning

Recent studies [2, 3, 4, 5, 11, 15, 16] have utilized deep learning techniques such as convolutional neural networks (CNNs) [17], multi-task learning networks, spatial transformer networks (STNs) [18], and transformer architectures such as Vision Transformers (ViT) [16] for improved performance in player identification and related tasks. Liu and Bhanu [4] proposed an approach that utilizes a variant of Region-based Convolutional Neural Networks (R-CNNs), which are well-known networks for object detection and localization tasks that exploit human body part cues for digit-level localization and classification. Vats et al. [11] used player 'tracklets' as input, employing a weakly supervised training methodology and leveraging frame-level labels to expedite training under a ViT-based architecture.

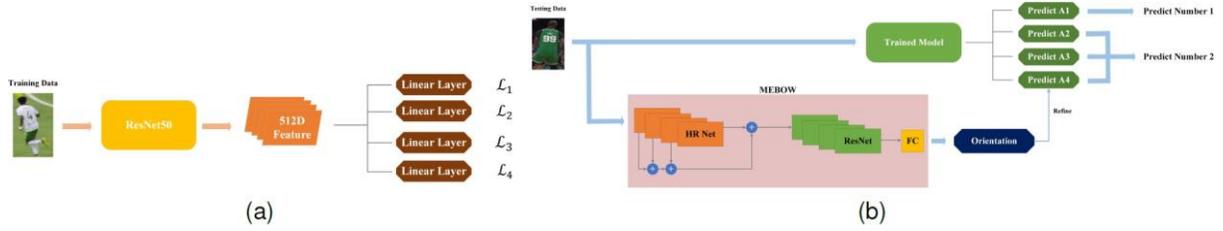

Figure 1: Architecture of the proposed system. (a) Training schematic diagram. We utilize ResNet50 as the backbone network to extract features. (b) Prediction schematic diagram with orientation-guided weight refinement.

### 2.3 JNR Using Training Approaches

Two output label representations can be used for training a network to recognize jersey numbers, namely, the holistic and digit-wise approaches [6]. The holistic approach considers an entire jersey number as one class. A network is then trained to predict jersey numbers as single entities, without distinguishing between individual digits. The digit-wise approach considers the two digits in a jersey number as two separate classes. In this approach, a network is trained to recognize each digit independently. Zhang et al. [9] proposed a new method to address the challenges of associating the detection and preservation of identities in player identification. Unlike existing approaches that rely on appearance models, their method proposes a distinct deep representation of player identity, improving accuracy despite the presence of similar player appearance within the same team. Recently, SoccerNet [19] has made a large amount of player video sequences available and developed a challenge for locating soccer events. SoccerNet-v2 [20], a new dataset based on the SoccerNet dataset, consists of approximately 300,000 annotations and 500 untrimmed soccer videos.

### 2.4 Pose-Guided Schemes in JNR

Researchers have extensively explored human pose analysis for JNR, mainly focusing on extracting skeletal structures and key points [21, 22, 23, 24, 25, 26, 27]. A noteworthy advancement highlighted in MEBOW [28] and [29] involved using pose estimation techniques to determine the rotational angles of the human body. Despite progress in this area, there is still a research gap regarding how these methods can be applied to JNR. Among the few studies that have explored this area, [4] and [8] stand out as major contributors.

Liu and Bhanu [4] used human pose information as a key method to prevent the erroneous capture of unrelated text found on sports fields. Their approach helps improve the accuracy and precision of JNR by ensuring that only relevant information is considered. In addition, their recent study [8] took a slightly divergent approach by leveraging player-specific features and pose estimations to guide the generation of more robust regions of interest. Despite the innovative strategies employed by these studies, it is important to note that neither study directly explored the intricate relationship between human poses and their influence on the recognition of numerical digits. They described a universal Jersey Number Detector (called JEDE) for player identification in sports analytics that predicts more robust proposals guided by player features and pose estimation. JEDE predicts players' bounding boxes and key points, along with the bounding boxes and classes of jersey digits and numbers, in an end-to-end manner. Ultimately, the successful accomplishment of such tasks is reliant upon the availability of a well-curated dataset.

## 3 Methodology

This study aims to explore new approaches in the field of JNR by utilizing multi-task learning as the primary framework. We adopt ResNet50 [30] as our backbone network to learn four different tasks: overall recognition, tens-digit recognition, ones-digit recognition, and digit number, thereby extracting 512-dimensional features. During the model training phase, we employ a cross-entropy loss function for parameter optimization and adjust the weights of each loss to enhance model performance. In addition, we introduce an angle-digit refine scheme (ADRS) to further optimize the predictive capability of the model. The remainder of this section describes a detailed explanation of our methodology and demonstrates the advantages of our approach through a comparison with other state-of-the-art methods. Furthermore, we have collected datasets from diverse team sports, including soccer, basketball, volleyball, baseball, and football, to ensure the generalization and practicality of the model.

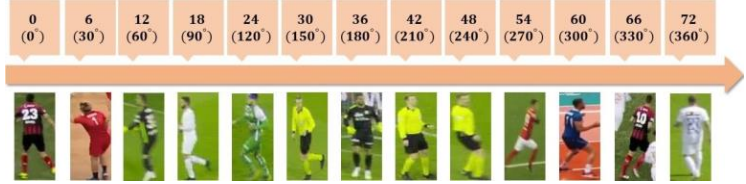

Figure 2: Orientation classification in this study.

### 3.1 Network Structure

We utilize multi-task machine learning as the kernel of our architecture, as illustrated in Figure 1. In the training phase, we employ ResNet50 as our backbone and learn four distinct tasks: overall recognition, tens-digit recognition, ones-digit recognition, and number of digits, as well as 512-dimensional features, all obtained from ResNet50.

These features are then passed through four linear layers, resulting in four probabilities: overall probability ($P_1$), probability of tens-digit numbers ($P_2$), probability of ones-digit numbers ($P_3$), and probability of the number of digits ($P_4$). Among these, $P_1 \in \mathbb{R}^{101}$ (representing jersey numbers 0-99, where 100 represents the category of invisible numbers), while $P_2$ and $P_3 \in \mathbb{R}^{11}$ (representing numbers 0-9, with 10 representing invisible numbers), and $P_4 \in \mathbb{R}^3$. Let $G_1$, $G_2$, $G_3$, and $G_4$ represent the ground truth values for overall recognition, tens-digit numbers, ones-digit numbers, and the number of digits, respectively. We utilize cross-entropy loss as our loss function, shown in equation 1, 2, 3, 4 and 5.

The loss function are defined as follows:

- Overall loss function:

$$\mathcal{L}_1 = -\sum_{i=0}^{100} G_1^i \log P_1 \quad (1)$$

- Tens-digit loss function:

$$\mathcal{L}_2 = -\sum_{j=0}^{10} G_2^j \log P_2 \quad (2)$$

- Ones-digit loss function:

$$\mathcal{L}_3 = -\sum_{k=0}^{10} G_3^k \log P_3 \quad (3)$$

- Number of digits loss function:

$$\mathcal{L}_4 = -\sum_{l=0}^{2} G_4^l \log P_4 \quad (4)$$

The total loss is given by

$$\mathcal{L}_{total} = \sum_{m=1}^{4} \alpha_m * \mathcal{L}_m$$

where $\alpha_1$, $\alpha_2$, $\alpha_3$ and $\alpha_4$ represent the weight of each loss, respectively.

At the prediction stage, we apply ADRS. For calculating human body rotation angle, we divide the angle into 5 degree intervals, with labels ranging from 0 to 72. Specific test results and a description of each angle interval are shown in Figure 2. In [28], HRNet [31] was utilized as a feature extractor, followed by ResNet and a softmax fully connected layer. After calculating the human body rotation angle, logical operations are performed on the angle according to Algorithm 1. We adjust certain parts of digits that may cause incorrect interpretations of numbers and digits due to excessive angles. This prevents the number of digits from being erroneously recognized as a wrong number due to excessive pose angle, thus obtaining incorrect output.

Table 1: Number of images for training, validation, and testing

| Training | Validation | Testing | Total |
|---|---|---|---|
| 45,099 | 8,490 | 4,974 | 58,563 |

**Algorithm 1** Number of digits with orientation

**Input:** Image, $A_2$, $A_4$, Orientation
**Output:** $A_4$

$A_2 \leftarrow$ *tens digit*
$0 \leq A_2 \leq 10$
$A_2 = 10$ represents that the tens digit of jersey number is invisible

$A_4 \leftarrow$ *number of digits*
$0 \leq A_4 \leq 3$
$A_4 = 0$ represents that the jersey number is invisible

$Orientation \leftarrow MEBOW(Image)$

**Procedure** DECISION ($A_2$, $A_4$, *Orientation*)
**if** $85° \leq Orientation \leq 265°$ and $A_2 \neq 10$ and $A_2 \neq 0$
    $A_4 \leftarrow 2$
**else**
    $A_4 \leftarrow A_4$
**end if**

**Algorithm 2** Top-1 prediction selection

**Input:** Image, $A_1$, $A_2$, $A_3$, $A_4$, $P_1$, $P_2$, $P_3$
**Output:** Predict
$A_1 \leftarrow$ *holistic jersey number*
$A_2 \leftarrow$ *tens digit*
$A_3 \leftarrow$ *ones digits*
$A_4 \leftarrow$ *number of digits*
$0 \leq A_4 \leq 3$
$A_4 = 0$ represents that the jersey number is invisible
$P_1 \leftarrow$ *the probability of* $A_1$
$P_2 \leftarrow$ *the probability of* $A_2$
$P_3 \leftarrow$ *the probability of* $A_3$
**Procedure** SELECTION ($A_1$, $A_2$, $A_3$, $A_4$, $P_1$, $P_2$, $P_3$)
**if** $A_4 = 2$
    $combined\_number \leftarrow int(str(A_2) + str(A_3))$
**else**
    $combined\_number \leftarrow A_3$
**end if**
**if** $P_1 > P_2$ and $P_1 > P_3$
    $Predict \leftarrow A_1$
**else**
    $Predict \leftarrow combined\_number$
**end if**
**if** $A_4 = 0$
    $Predict = 100$
**end if**

The ADRS setting in this article is configured from 85 degrees to 265 degrees, ensuring smooth output for the predictions. Our architecture can produce two types of predictions: Prediction Number 1 (via overall prediction) and Prediction Number 2 (through individual number learning and ADRS), as shown in Figure 1(b). If either prediction type is accurate, we consider the prediction to be correct. We refer to this as Top-2 grading. Additionally, we apply Predict Selection, outlined in Algorithm 2, which involves predicting the original two outputs and applying probability to identify a relatively accurate result as the final prediction. This grading approach is referred to as Top-1.

### 3.2 Dataset

The dataset used in this study combines various team sport data, including soccer, football, basketball, volleyball, and baseball data. Our approach differs from [6], which solely focused on ice hockey images. We primarily utilize the SoccerNet dataset to select relevant data. In addition, we combine videos from multiple games to extract suitable jersey number data. As depicted in Figure 3, our dataset comprises a total of 58,563 images, with the training set comprising 45,099 images, the validation set containing 8,490 images, and the test set consisting of 4,974 images, as shown in Table 1. The composition of the dataset consists of 42,500 images of soccer (96% of the total), 666 images of football (2% of the total), 247 images of baseball (1% of the total), 522 images of basketball (1% of the total) and 1,095 images of volleyball (2% of the total). Although the overall percentage of sports excluding soccer is relatively low (6%), these data provide crucial support for rare jersey numbers (numbers 60-99). The dataset accounts for approximately 34% of these less popular jersey numbers, as illustrated in Figure 4. These data play a significant role in supplementing and completing our dataset.

## 4 Experiments

We identified a optimal backbone network and determined the most effective loss weights through two ablation studies. The first set of studies examined the influence of various backbone networks and batch sizes on accuracy, aiming to find the backbone network with the highest accuracy and the most appropriate batch size. The second set of studies evaluated the impact of different loss weights on accuracy to determine the best combination. Through these studies, our goal was to enhance model performance, improve the accuracy of predicting jersey numbers, and identify the best training parameter settings.

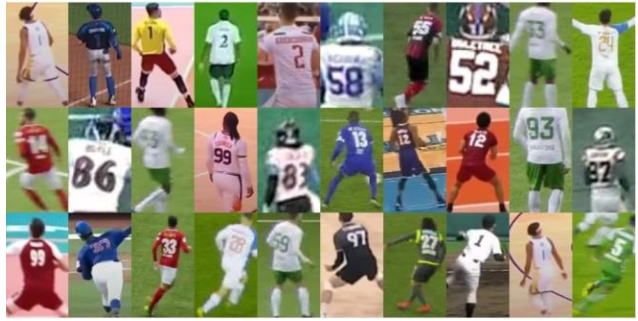

Figure 3: Sample images in the dataset. Our dataset integrates a variety of team sports such as soccer, baseball, basketball, volleyball, and football.

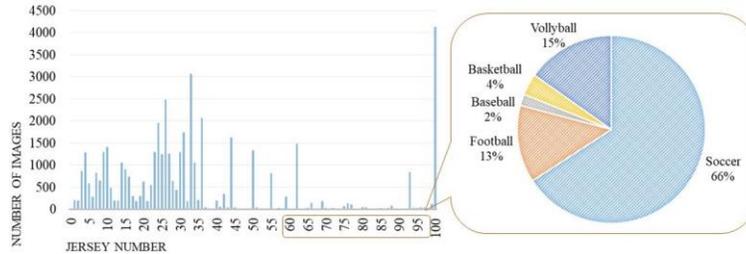

Figure 4: Number of images and exercise distribution for rare jersey numbers in the dataset (number 60-99).

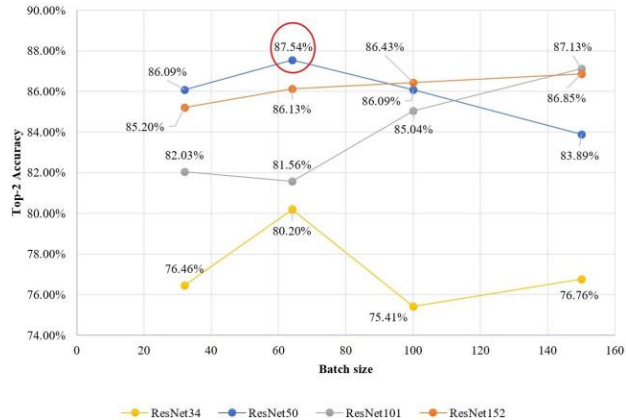

Figure 5: Results of ablation study with respect to backbone network and batch size.

### 4.1 Ablation Study

To find the optimal backbone and best loss weights [32], we conducted two different ablation studies. The first set of studies assessed the accuracy of different backbones and batch sizes under consistent conditions of epoch, loss weight, and optimizer to understand their impact on accuracy. Our goal was to identify the backbone with the highest accuracy and determine the most suitable batch size. With the epoch set to 100, loss weights ($\alpha_1$, $\alpha_2$, $\alpha_3$, and $\alpha_4$) set to 1, and using the Adam optimizer [33], we observed that, for ResNet34 and ResNet50, the highest accuracy was achieved with a batch size of 64. For network models with larger architectures like ResNet101 and ResNet152, the batch size yielding the highest accuracy was 150. Figure 5 illustrates that the highest accuracy was attained using ResNet50 with a batch size of 64.

Table 2: Counts of parameters and FLOPs (Floating Point Operations per Second) of the backbone network and batch size in the ablation study

| Model | Batch size | FLOP | Parameters |
|---|---|---|---|
| ResNet34 | 32 | 4.8 G | 21.3 M |
| | 64 | 9.6 G | |
| | 100 | 15.0 G | |
| | 150 | 22.5 G | |
| ResNet50 | 32 | 5.4 G | 23.8 M |
| | 64 | 10.8 G | |
| | 100 | 16.9 G | |
| | 150 | 25.4 G | |
| ResNet101 | 32 | 10.3 G | 42.8 M |
| | 64 | 20.6 G | |
| | 100 | 32.2 G | |
| | 150 | 48.2 G | |
| ResNet152 | 32 | 15.2 G | 58.4 M |
| | 64 | 30.3 G | |
| | 100 | 47.4 G | |
| | 150 | 71.1 G | |

Table 3: Comparison of JNR with various methods and configurations

| | Method | Accuracy | Precision | Recall | F1 score |
|---|---|---|---|---|---|
| Top-1 | Vats[6] (our dataset) | 53.26% | 71.09% | 53.26% | 56.58% |
| | **Ours** | **64.07%** | **73.64%** | **64.07%** | **67.46%** |
| | **Compare to [6]** | **(+10.81%)** | **(+2.55%)** | **(+10.81%)** | **(+10.88%)** |
| | Baseline (Random) | 0.99% | 12.15% | 0.99% | 28.84% |
| Top-2 | Vats[6] (our dataset) | 78.59% | 89.39% | 78.59% | 79.54% |
| | **Ours** | **89.97%** | **91.33%** | **89.96%** | **90.64%** |
| | **Compare to [6]** | **(+11.38%)** | **(+1.94%)** | **(+11.37%)** | **(+11.10%)** |
| | Baseline (Random) | 1.83% | 14.88% | 1.83% | 16.69% |

Table 4: Quantitative results of the ablation study with respect to loss weights

| $\alpha_1$ | $\alpha_2$ | $\alpha_3$ | $\alpha_4$ | Accuracy |
|---|---|---|---|---|
| 0.25 | 0.25 | 0.25 | 0.25 | 88.50% |
| 0.3 | 0.25 | 0.25 | 0.2 | 88.70% |
| **0.2** | **0.3** | **0.3** | **0.2** | **89.97 %** |
| 0.1 | 0.4 | 0.4 | 0.1 | 87.60 % |
| 0.4 | 0.25 | 0.25 | 0.1 | 88.02% |

We also calculated the parameters and FLOPs for each combination in the ablation studies, as shown in Table 2. We found that using ResNet50 instead of ResNet34 as the backbone resulted in no significant difference in parameter growth and FLOPs. However, when ResNet101 or ResNet152 was utilized as the backbone, there was a noticeable increase in both parameter and FLOP growth, as well as lower accuracy compared to ResNet50. Therefore, we concluded that ResNet50 with a batch size of 64 was the optimal choice for our model.

The second set of ablation studies tested the impact of different loss weights on the accuracy rate under consistent conditions of epoch, optimizer, backbone, and batch size. Our goal was to identify the combination of loss weights that yields the highest accuracy. We set the epoch to 100, utilized the Adam optimizer, and employed the best backbone and batch size combination found from the previous experiment (ResNet50 with a batch size of 64). We tested 5 different combinations and found that when $\alpha_1$=0.2, $\alpha_2$=0.3, $\alpha_3$=0.3, and $\alpha_4$=0.2, the highest accuracy was achieved, reaching 89.97% accuracy, 91.33% precision, 89.96% recall, and 90.64% F1 score, as shown in Tables 3 and 4.

### 4.2 Result

We compared the results of our architecture with that of [6] using our dataset, utilizing randomly generated predictions as a baseline. Referring to Table 3, under the same conditions of epoch (epoch = 100), batch size (batch size = 64), and backbone (ResNet50), and using the optimal loss weights suitable for each condition, our accuracy improved upon [6] by 10.81% in Top-1 and 11.38% in Top-2, with the F1 scores also improved by 10.88% and 11.10%, respectively. These results indicate that our architecture outperforms a method that solely divides the task into overall and individual numbers. The inclusion of ADRS enhances our model's ability to accurately predict jersey numbers. Another potential factor contributing to this improvement is that [6] only considers a single sport (i.e., ice hockey) and utilizes only 81 categories, making the problem relatively simpler. While a simpler architecture may achieve higher accuracy for a less complex problem, its accuracy would be expected to decrease when applied to jersey numbers across various sports.

Table 5: Comparison of dataset image counts

| Method | #Images | #Classifications |
|---|---|---|
| Vats[6] | 54,251 | 81 |
| Gerke[5] | 8,281 | 36 |
| Liu[4] | 3,567 | 100 |
| **Ours** | **58,563** | **101** |

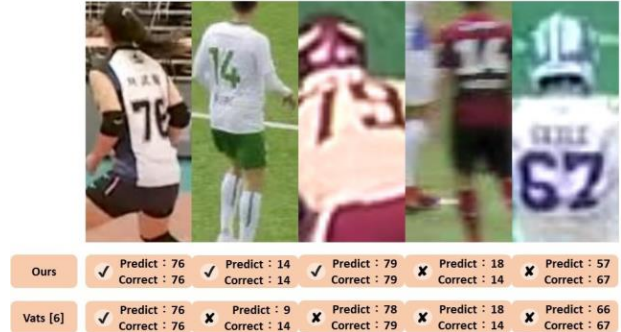

Figure 6: Examples of system robustness analysis.

Our dataset contains a greater number of images than other studies. As indicated in Table 5, the number of images used by Liu et al. [4] and Gerke et al. [5] was 3,567 and 8,281 images, respectively. While Vats et al.'s [6] dataset comprised 54,251 images, our dataset has 58,563. Having more data allows our model to assimilate a greater amount of digital information.

Regarding the number of categories, [5] uses only 36 classes, [6] uses 81 categories, and, although [4] has 100 classes, it separates ones-digits and tens-digit for recognition, distinguishing which number from 0 to 9 each represents. This implies that it does not consider them as a single entity. Therefore, our method also uses the highest number of categories.

### 4.3 System Robustness Analysis

Here, we cover some of the visual and recognition results in Figure 6. In the first image, although there is a significant rotation of the human body, both our method and Vats et al. [6] correctly determine the jersey. In the second image, the advantage of using our method is evident. Due to the rotation of the body, the tens-digit '1' might be incorrectly interpreted as non-existent. Our method successfully predicts it as a two-digit number and predicts the number correctly, whereas [6] predicts it as a single digit number. The third image is a considerably motion-burred image, and the player is bending forward (the angle of the letters tilts forward). Our method can also make correct judgments in this case. In the fourth image, due to the extreme blurriness of the two digits and the appearance of a second person in the image, neither our method nor [6] is able to obtain the correct answer. The fifth image has wrinkles in the clothing and blurred characters, leading to incorrect recognition of the tens-digit both by our method and that of [6].

## 5 Conclusion and Future Work

This paper presents an innovative multi-task architecture for JNR that learns the overall number, tens-digit, ones-digit, and number of digits. We introduce an angle-digit refine scheme (ADRS) at the prediction stage to help adjust digit positions based on human body rotation angles. Our dataset encompasses various sports, enhancing the practicality of the model. It consists of 101 categories (the numbers 0-99 and an additional category representing invisible numbers). The inclusion of digit information significantly improves our model accuracy compared to other multi-task architectures. We also propose Top-1 and Top-2 prediction modes, which broaden the application scenarios.

In future work, we will address the issue of class imbalance in our dataset. Due to difficulties in collecting images for certain less popular jersey numbers, we will continuously update our dataset to make it more comprehensive and balanced.